\documentclass[runningheads]{llncs}
\usepackage{graphicx}
\usepackage{gensymb}
\usepackage{amsmath}
\usepackage{hyperref}
\usepackage[tickmarkheight=1pt]{todonotes}
\begin{document}
%
\title{Need for Design Patterns: Interoperability Issues and Modelling Challenges for Observational Data}
%
\author{Trupti Padiya\inst{1,3}\orcidID{0000-0003-4514-4473} \and
	Frank Löffler\inst{1,2}\orcidID{0000-0001-6643-6323} \and
	Friederike Klan\inst{2,3}\orcidID{0000-0002-1856-7334}}
\authorrunning{T.Padiya et al.}
\titlerunning{Need for Design Patterns for Observational Data}
%
\institute{Heinz Nixdorf Chair for Distributed Information Systems, Friedrich Schiller University Jena, Germany \and
	Competence Center for Digital Research, Michael Stifel Center Jena, Friedrich Schiller University Jena, Germany \and
	Institute of Data Science, German Aerospace Center (DLR), Jena, Germany
	\email{\{trupti.padiya,frank.loeffler\}@uni-jena.de}\\
	\email{\{trupti.padiya,friederike.klan\}@dlr.de}}
\maketitle              
%

\begin{abstract}
Interoperability issues concerning observational data have gained attention in recent times. Automated data integration is important when it comes to the scientific analysis of observational data from different sources. However, it is hampered by various data interoperability issues. We focus exclusively on semantic interoperability issues for observational characteristics.

We propose a use-case driven approach to identify general classes of interoperability issues. In this paper, this is exemplarily done for the use-case of citizen science fireball observations. We derive key concepts for the identified interoperability issues that are generalizable to observational data in other fields of science. These key concepts contain several modeling challenges and we broadly describe each modeling challenge associated with its interoperability issue. We believe, addressing these challenges with a set of ontology design patterns will be an effective means for unified semantic modeling, paving the way for a unified approach for resolving interoperability issues in observational data. We demonstrate this with one design pattern, highlighting the importance and need for ontology design patterns for observational data, and leave the remaining patterns to future work. Our paper thus describes interoperability issues along with modeling challenges as a starting point for developing a set of extensible and reusable design patterns.

\keywords{Citizen Science  \and Fireball observations \and Observational data \and Ontology design patterns \and Semantic interoperability.}
\end{abstract}
\footnote{This work was carried out when the first author was affiliated with Friedrich Schiller University Jena, Germany, and Institute of Data Science, DLR Jena, Germany. At the time of submission on arxiv, the first author is affiliated with
TIB Leibniz Information Centre for Science and Technology, Hanover, Germany.}
\section{Introduction}
Interoperability issues with regard to observational data have recently gained attention~\cite{RDA}. Observational data can have different formats, structures, and diversity in semantic representation of observational characteristics, resulting in interoperability issues at all these levels. In this paper, we focus exclusively on issues regarding data interoperability in the context of semantic representations of observational characteristics. Automated data integration is vital for observational data. However, this is hard to achieve. One of the major reasons is that use-cases and data interoperability are not taken into consideration during the development of observational data models. As a result, a variety of observational data models came gradually into existence and this diversity in data modeling approaches is hampering automated data integration~\cite{schindler2019automated}.

In order to tackle this fundamental problem, we propose a bottom-up approach which is a use-case driven approach. Based on real-world requirements and discussions with domain experts, we identify semantic interoperability issues exemplarily for the use-case of citizen science fireball observations. We derive key concepts for the identified interoperability issues that are generalizable to observational data in other fields of science. The ability to derive key concepts for the identified issues being applicable to general observational data demonstrates the potential of our approach. These derived key concepts contain several modeling challenges which we broadly describe. We believe, these challenges can be resolved in a unified and reusable manner by developing corresponding ontology design patterns. Moreover, existing design patterns can foster reuse of best practice solutions for modeling, resulting in more robust and unified modeling solutions. Here, we limit ourselves to a demonstration of one such design pattern. We call for the community to follow this approach and contribute ontology design patterns that can address interoperability issues for observational data, especially for citizen science.

\section{Use-Case: Citizen Science Fireball Observations}
Citizen science fireball observations consist of vital information about fireball-observations and are used as a basis for scientific study and analysis of fireballs and fireball events, e.g. determining the trajectory and mass of the observed objects. Typically, a fireball observation can be described as follows: \textit{A fireball event occurs in the sky, and an observer witnesses the event and records various relevant properties, e.g., duration of event and color of the fireball. These observational data are then reported through citizen science reporting platforms, usually via web-app or web-form.} There are several citizen science apps to report fireball observations like Skywatcher from Nachtlicht-BüHNE~\cite{skywatcher}, Fireballs in the sky~\cite{fireballSky} and many others. Web-forms are also provided from different fireball networks across the world, like AMS/IMO~\cite{IMO}, DFN~\cite{DFN}, Allsky7~\cite{AllSky7}, ESA~\cite{ESA}, and many others. Our use-case is based on real-world datasets from AMS/IMO, Nachtlicht-BüHNE, data from camera images, and data derived by scientists. We also consider concrete data needs from scientists for identifying and modeling interoperability issues inherent to fireball observational data.
\begin{table}
\centering
\caption{Example data from different citizen science fireball observational datasets pertaining to same semantics}\label{tab1}
\resizebox{\columnwidth}{!}{%
\begin{tabular}{|l|l|l|l|}
\hline
Observational properties & Dataset1 &  Dataset2 & Dataset3\\
\hline
location & Jena, Germany & 07745, Jena & 50.9271\degree N, 11.5892\degree E \\
level of experience & intermediate & 7 & good \\
viewing-direction & 57\degree & north & east-northeast\\
color & red & yellow, blue, white &	\#FF0000,\#E6E6FA\\
fireball-train-duration & 0.033 minute & unknown & 1 second \\
moving-direction & 112.5\degree & east to southeast & top-left to bottom-right\\
\hline
\end{tabular}%
}
\end{table}

Scientific analysis of these data calls for data integration from multiple fireball observation datasets. However, there are several interoperability challenges.  e.g.,~observable properties of the fireball are represented differently in different datasets, both at schema level and instance level, despite having the same semantics. Table \ref{tab1} depicts data at instance level. It is important to map these semantically similar data in order to get an accurately integrated dataset.  

\section{Semantic Interoperability Issues and Modelling Challenges}
We describe some of the semantic interoperability issues along with modeling challenges based on our use-case. We derive key concepts that represent the generalizable interoperability issue, which eminently apply to observational data in other fields of science. Derived key concepts are written as a header for every challenge. Furthermore, we demonstrate one design model to resolve one of the listed challenges.

\textbf{\textit{Expertise-Level: }} Table~\ref{tab1} shows various ways of representing the level of experience in our use-case. It is represented using an interval scale with varying ranges like $[1-5]$ or $[1-3]$ or an ordinal scale referring to non-mathematical ideas like (beginner, intermediate, expert) or (poor, fair, average, good, excellent). There is a possibility of mapping interval scale to ordinal scales whereas doing the reverse is problematic. Moreover, it is important to map varying ranges that belong to a similar scale of measure. Semantically mapping different representations of expertise-level is a general interoperability issue that needs to be addressed.

\textbf{\textit{Location: }}Table~\ref{tab1} also depicts some of the ways a location of observer is represented in our use-case. In itself, it is a general interoperability issue, as location is represented differently by location name, a postal code, nearby location, or geo-coordinates as latitude and longitude. Furthermore, latitude/longitude can also be represented as ecliptic, galactic, and supergalactic latitude/longitude. Sometimes, latitude and longitude are represented with a notion of declination and right ascension respectively. Additionally, different units and formats exist to represent the coordinates.

\textbf{\textit{Visual characteristics: }} In our use-case, the property color is represented in different ways as shown in Table~\ref{tab1}. Color is represented by dictionary color names or color codes like RGB, HSV, and others. It is possible to map dictionary color names to color codes. However, not all color codes have names. Moreover, there are color names that are difficult to represent with a color code, e.g.,~light blue-green. In addition, a fireball or other feature might show temporal variation of color. E.g., a sequence of colors appear with a timeline and it is important to maintain this information while integrating data. Same challenges are applicable to a large fraction of observable objects, e.g., sky, animals, plants. Here different semantic representations of visual characteristics is a generalizable interoperability issue that covers not only color but also shape, form, texture, etc.

\textbf{\textit{Relative and cardinal directions: }}The moving direction of a fireball can have multiple representations as shown in Table \ref{tab1}. It may be represented using angular degrees, or body relative directions like left, right, up (top), down (bottom), forward, backward, and their permutations like left to right, or top-left to top-right, and so on with varying granularity information. Another representation is using cardinal directions like east to west, or northeast to northwest, or north-northeast to north-northwest, and so on with different granularity information. The generalizable interoperability issue is mapping angles, relative directions, and cardinal directions with additional granularity information.


\textbf{\textit{Angles and directions: }}In our use-case, viewing-direction refers to the direction that an observer is facing while observing an event. It is represented either using a cardinal direction or a viewing angle. Moreover, the granularity level of the cardinal direction also varies in different datasets. E.g., some datasets might have 8 compass sectors: east (E), west(W), north(N), south(S), north-east(NE), north-west(NW), south-east(SE), south-west(SW). Whereas some might have 16 compass sectors: north northeast (NNE), east northeast(ENE), and so on in addition to 8 compass sectors. One such example of variation in granularity and representation in the actual dataset is shown in Table \ref{tab1}.

\begin{figure}
    \centering
    \includegraphics[scale=0.90]{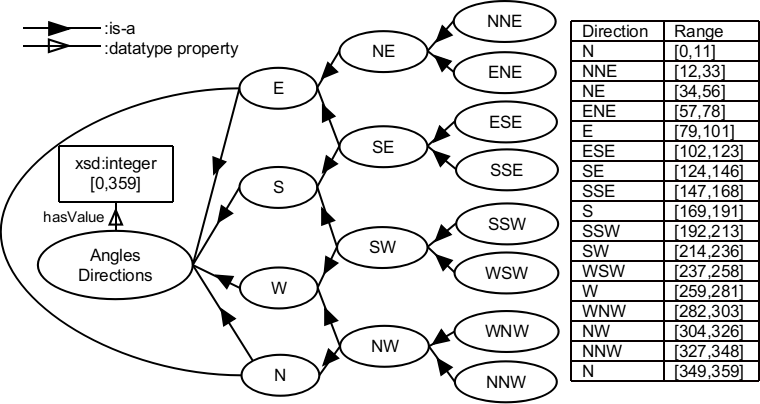}
    \caption{Graphical representation of modeling: Angles and Directions}
    \label{fig1}
\end{figure}

A solution to the general interoperability issue of mapping different concepts like angles and directions with appropriate granularity information is depicted in Fig. \ref{fig1}. We use SWRL rules to infer angles from directions and vice-versa. An example of SWRL rule for NNE is: $ AnglesDirections (?v) \wedge hasValue(?v,?x) \wedge swrlb:greaterThan (?x,12) \wedge swrlb:lessThan (?x,33) \implies NNE (?v) $.\newline
Similarly, for every direction, SWRL rules are created using the ranges as given in Fig. \ref{fig1}.  This modeling solution can be reused, e.g., in wind observations, and/or can be extended for other domains based on requirements.

The modeling solution in Fig. \ref{fig1} is inspired by the view-inheritance design pattern~\cite{ODPinher}, which demonstrates that extensible design patterns foster re-usability, provide clear modeling structure,  decrease complexity, increase reliability, and become robust with progression, which gradually leads towards unified modeling solutions. We believe, addressing modeling challenges in this manner will be an effective way to unified semantic modeling of interoperability aspects and thus enabling automatic data integration. It is important to note that modeling observational data in a unified way is a prerequisite for resolving interoperability issues in a unified manner using design patterns. Moreover, there is a strong requisite of design patterns for observational data and their characteristics.  

\section{Conclusions}
In this paper, we describe some of the important semantic interoperability challenges for observational data. We argue that addressing these challenges in terms of a set of extensible and reusable ontology design patterns will be an effective way to unified semantic modeling of interoperability aspects leading towards automatic semantic data integration. We demonstrated a potential way to address semantic interoperability issues in a unified manner. It will accommodate for other scientific domains that leverage observational data to reuse these design patterns and extend them based on their needs. It is a call for the community to address modeling challenges by contributing ontology design patterns.


\bibliographystyle{splncs04}
\bibliography{ref.bib}

\end{document}